\documentclass{article}

\usepackage{microtype}
\usepackage{graphicx}
\usepackage{subfigure}
\usepackage{booktabs} % for professional tables
\usepackage{multirow}
\usepackage{multicol}

\usepackage{hyperref}

\usepackage[accepted]{icml2025}
\usepackage{icml2025}

\usepackage{amsmath}
\usepackage{amssymb}
\usepackage{mathtools}
\usepackage{amsthm}

\usepackage[capitalize,noabbrev]{cleveref}

\theoremstyle{plain}

\theoremstyle{definition}

\theoremstyle{remark}

\usepackage[textsize=tiny]{todonotes}

\begin{document}

\twocolumn[
\icmltitle{Actionable Interpretability via Causal Hypergraphs: Unravelling Batch Size Effects in Deep Learning
 }

% \\ International Conference on Machine Learning (ICML 2025)



% \begin{icmlauthorlist}
% \icmlauthor{Zhongtian Sun}{yyy}
% \icmlauthor{Anoushka Harit}{sch}
% \icmlauthor{Pietro Lio}{sch}
% % \icmlauthor{Firstname4 Lastname4}{sch}
% % \icmlauthor{Firstname5 Lastname5}{yyy}
% % \icmlauthor{Firstname6 Lastname6}{sch}
% % \icmlauthor{Firstname7 Lastname7}{comp}
% % %\icmlauthor{}{sch}
% % \icmlauthor{Firstname8 Lastname8}{sch}
% % \icmlauthor{Firstname8 Lastname8}{yyy,comp}
% \end{icmlauthorlist}

% \icmlaffiliation{yyy}{University of Kent}
% \icmlaffiliation{sch}{University of Cambridge}

\begin{center}
\textbf{Zhongtian Sun}$^{1}$, \textbf{Anoushka Harit}$^{2}$, \textbf{Pietro Liò}$^{2}$ \\
$^{1}$University of Kent \quad
$^{2}$University of Cambridge
\end{center}


\icmlkeywords{Machine Learning, ICML}

\vskip 0.3in
]



% \printAffiliationsAnd{}

\begin{abstract}
While batch size’s impact on generalisation is well-studied in vision tasks, its causal mechanisms remain underexplored in graph and text domains. We introduce a hypergraph-based causal framework, HGCNet, that leverages deep structural causal models (DSCMs) to uncover how batch size influences generalisation via gradient noise, minima sharpness, and model complexity. Unlike prior approaches based on static pairwise dependencies, HGCNet employs hypergraphs to capture higher-order interactions across training dynamics. Using do-calculus, we quantify direct and mediated effects of batch size interventions, providing interpretable, causally grounded insights into optimisation. Experiments on citation networks, biomedical text, and e-commerce reviews show HGCNet outperforms strong baselines including GCN, GAT, PI-GNN, BERT, and RoBERTa. Our analysis reveals that smaller batch sizes causally enhance generalisation through increased stochasticity and flatter minima, offering actionable interpretability to guide training strategy in deep learning. This work positions interpretability as a driver of principled architectural and optimisation choices beyond post hoc analysis.
\end{abstract}

\section{Introduction}
Generalisation in deep learning is shaped by architectural choices and optimisation hyperparameters. Among these, batch size exerts a critical influence by modulating gradient noise, minima sharpness, and model complexity \cite{zhang2021understanding}. While extensively analysed in image classification \cite{keskar2016large, smith2018don}, batch size effects remain underexplored in graph and text domains, where dependencies are more intricate and existing theories do not fully transfer.

To address this gap, we introduce \textbf{HGCNet}, a hypergraph-based causal framework that extends deep structural causal models (DSCMs) to analyse batch size effects on generalisation. Unlike standard causal graphs, which capture only pairwise interactions, HGCNet models higher-order dependencies among training variables, enabling a principled understanding of how batch size influences generalisation via gradient noise, minima sharpness, and complexity. Using do-calculus \cite{pearl2009causality}, we quantify both direct and mediated causal effects of batch size interventions.

We evaluate HGCNet across diverse domains citation networks (Cora, CiteSeer \cite{sen2008collective}), biomedical text (PubMed \cite{pubmed2016}), and e-commerce reviews (Amazon \cite{mcauley2013hidden}) using both graph neural networks (GCN \cite{kipf2016semi}, GAT \cite{velickovic2017graph}, GIN \cite{xu2018powerful}, PI-GNN \cite{liu2023physics}, BGAT \cite{sun2021generative}) and transformer-based models (BERT \cite{devlin2019bert}, RoBERTa \cite{liu2019roberta}, DistilBERT \cite{sanh2019distilbert}). Our findings reveal that smaller batch sizes causally promote flatter minima and increased gradient noise, leading to consistent improvements in generalisation.

\begin{figure}[htbp]
    \centering
    \includegraphics[width=1.1\linewidth]{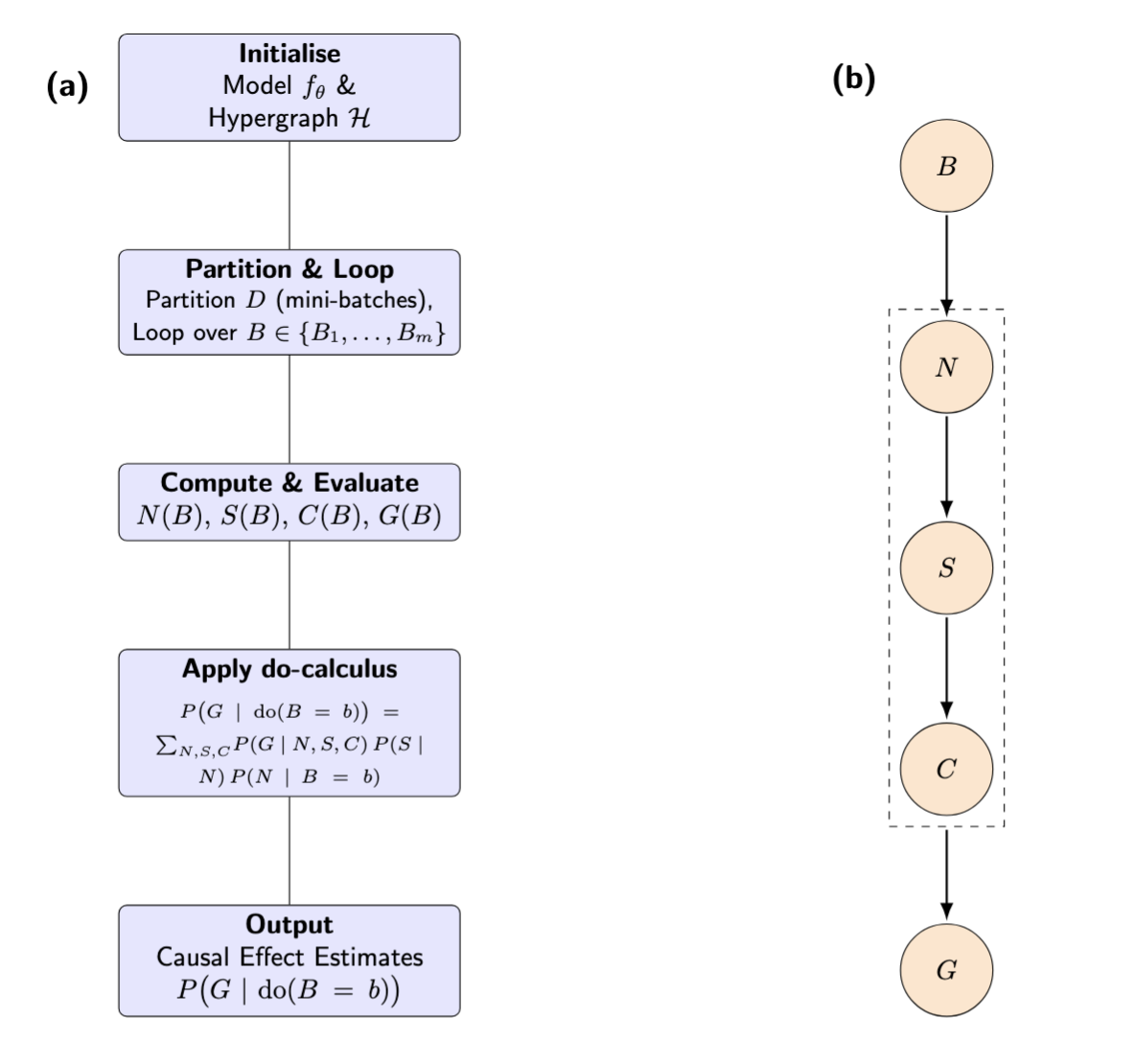}
    \caption{\textbf{(a)} Overview of the causal inference procedure. \textbf{(b)} Causal hypergraph with $B \rightarrow N \rightarrow S \rightarrow C$ and joint influence of $\{N, S, C\}$ on $G$.}
    \label{fig:1}
\end{figure}

\textbf{Our contributions are:}
\begin{enumerate}
    \item \textbf{Causal Hypergraph Framework:} We present HGCNet, a novel causal hypergraph model capturing higher-order dependencies in optimisation dynamics.
    \item \textbf{DSCM-Based Batch Analysis:} We formulate a principled DSCM to trace the influence of batch size through gradient noise, sharpness, and complexity.
    \item \textbf{Causality-Guided Optimisation:} We apply do-calculus to estimate actionable causal effects, supporting interpretable hyperparameter tuning.
    \item \textbf{Cross-Domain Validation:} We demonstrate consistent generalisation gains (2–4\%) across graph and text domains, validating the universality of the learned causal pathways.
\end{enumerate}
Our results offer both theoretical insight and practical guidance, positioning batch size as an interpretable and controllable lever in deep learning. The overall framework is shown in Figure~\ref{fig:1}, with methodology detailed in Section~\ref{method}.

\section{Related Work}
Batch size plays a crucial role in determining a model's ability to generalize in deep learning. \cite{keskar2016large} found that larger batch sizes often lead to convergence at sharp minima in the loss landscape, which negatively impacts generalisation.\cite{smith2018disciplined} extended this by demonstrating that smaller batch sizes introduce higher gradient noise, helping models escape sharp minima and converge to flatter regions, leading to better generalization in images. They observed that batch sizes smaller than 32 consistently yielded improved test performance by promoting a broader exploration of the loss landscape.

In addition,\cite{mertikopoulos2020almost} formalised the gradient noise hypothesis, showing that smaller batch sizes mitigate overfitting by preventing convergence to sharp minima. This aligns with \cite{wilson2017marginal}, who found that stochastic updates from small batch training enhance generalization by avoiding overfitting prone sharp minima. Similarly, \cite{dinh2017sharp} and \cite{hoffer2017train} demonstrated that flatter minima improve robustness, making models less sensitive to input variations.

While these findings provide strong insights for image classification, the effects of batch size on generalization in graph-based and text-based tasks remain underexplored.\cite{zhang2020deep} suggested that smaller batch sizes enhance robustness in Graph Convolutional Networks (GCNs)\cite{kipf2016semi} by facilitating better exploration of the solution space, but they did not conduct a rigorous causal analysis to establish a definitive relationship. In the text domain, \cite{radford2019language} examined the effects of batch size on learning stability in transformer-based models such as BERT but lacked a systematic causal investigation into the influence of batch size on generalization.

Existing work primarily focuses on the correlation between batch size and generalization but does not systematically formalize causal pathways linking batch size to key generalization factors such as gradient noise, minima sharpness, and model complexity. While recent advances in causal inference \cite{lin2021generative} have quantified the effects of hyperparameters on model performance, these methods have not been applied to analyze batch size effects in graph-based and transformer-based architectures.

To bridge this gap, we introduce HGCNet method that leverages Deep Structural Causal Models (DSCMs) and do-calculus to quantify the direct and mediated effects of batch size on generalization. Our approach explicitly models how batch size influences gradient noise, minima sharpness, and model complexity, providing a principled framework for understanding and optimizing batch size selection across diverse tasks, including graph and text classification.

\section{Problem Formulation}
The impact of batch size $(B)$ on generalization $(G)$ has been extensively studied in image classification tasks, but its effects on graph-based tasks (e.g., node classification, link prediction) and text classification tasks (e.g., sentiment analysis, document classification) remain underexplored. This work aims to analyze the causal pathways through which $B$ influences $G$, mediated by gradient noise $(N)$, minima sharpness $(S)$, and model complexity $(C)$. We introduce a causal hypergraph framework to model these relationships, capturing higher-order dependencies and multi-variable interactions that traditional causal graphs cannot represent.

Smaller batch sizes increase gradient noise $(N)$, which enhances exploration of the loss landscape and leads to flatter minima $(S)$. Flatter minima, in turn, reduce model complexity $(C)$, improving generalization $(G)$. These relationships are formalized as:

\begin{equation} \label{eq1}
N(B) \propto \frac{1}{B}, \quad S(B) \propto \frac{1}{B}, \quad G(S) \propto \frac{1}{S}
\end{equation}

Our framework disentangles the direct and mediated effects of $B$ on $G$, offering a structured approach to analyze and optimize batch size for enhanced generalization across diverse tasks. By leveraging the causal hypergraph, we address the limitations of traditional causal graphs in representing complex, multi-variable interactions, providing a more nuanced understanding of training dynamics in graph-based and text classification models.

\section{Methodology}
\label{method}
We formulate a causal framework to analyse how batch size influences generalisation through its downstream effects on gradient noise, minima sharpness, and model complexity. Traditional causal graphs often miss multi-variable interactions critical to optimisation. Therefore, we adopt a hypergraph-based Deep Structural Causal Model (DSCM) that captures higher-order dependencies and enables principled causal estimation using do-calculus.

\subsection{Causal Hypergraph Construction}
We represent the system using a causal hypergraph $\mathcal{H}=(\mathcal{V}, \mathcal{E})$, where:

$V=\{B, N, S,C, G\}$ represents the set of variables:
\begin{itemize}
    \item $B$ : Batch size (treated as the intervention variable)
    \item $N$ : Gradient noise (introduced due to mini-batching)
    \item $S$ : Minima sharpness (related to the curvature of the loss function)
    \item $C$ : Model complexity (a measure of hypothesis space size).
    \item $G$ : Generalization (measured as test accuracy)
\end{itemize}

The $\mathcal{E}$ is the set of directed edges representing causal dependencies:
\begin{itemize}
    \item $B \rightarrow N$ : Batch size affects the gradient noise due to the stochastic nature of gradient estimates in mini-batch training.
    \item $N \rightarrow S$ : The gradient noise influences the sharpness of the minima.
    \item $\{N, S\} \rightarrow C$ : Joint influence of gradient noise and minima sharpness on model complexity.
    \item $C \rightarrow G$ : Model complexity impacts generalization.
\end{itemize}
The hypergraph representation captures the indirect and mediated effects of $B$ on $G$, expressed as:

\begin{equation}
B \rightarrow N \rightarrow S \rightarrow C \rightarrow G
\end{equation}
Additionally, the edge $\{N, S\} \rightarrow C$ explicitly models the combined effect of gradient noise and minima sharpness on model complexity. While the causal hypergraph assumes linear or approximately linear relationships between variables, this simplification is justified by both theoretical and empirical evidence \cite{peters2017elements, zhang2021understanding}. 

\begin{figure}[ht]
    \centering
    \includegraphics[width=0.40\linewidth]{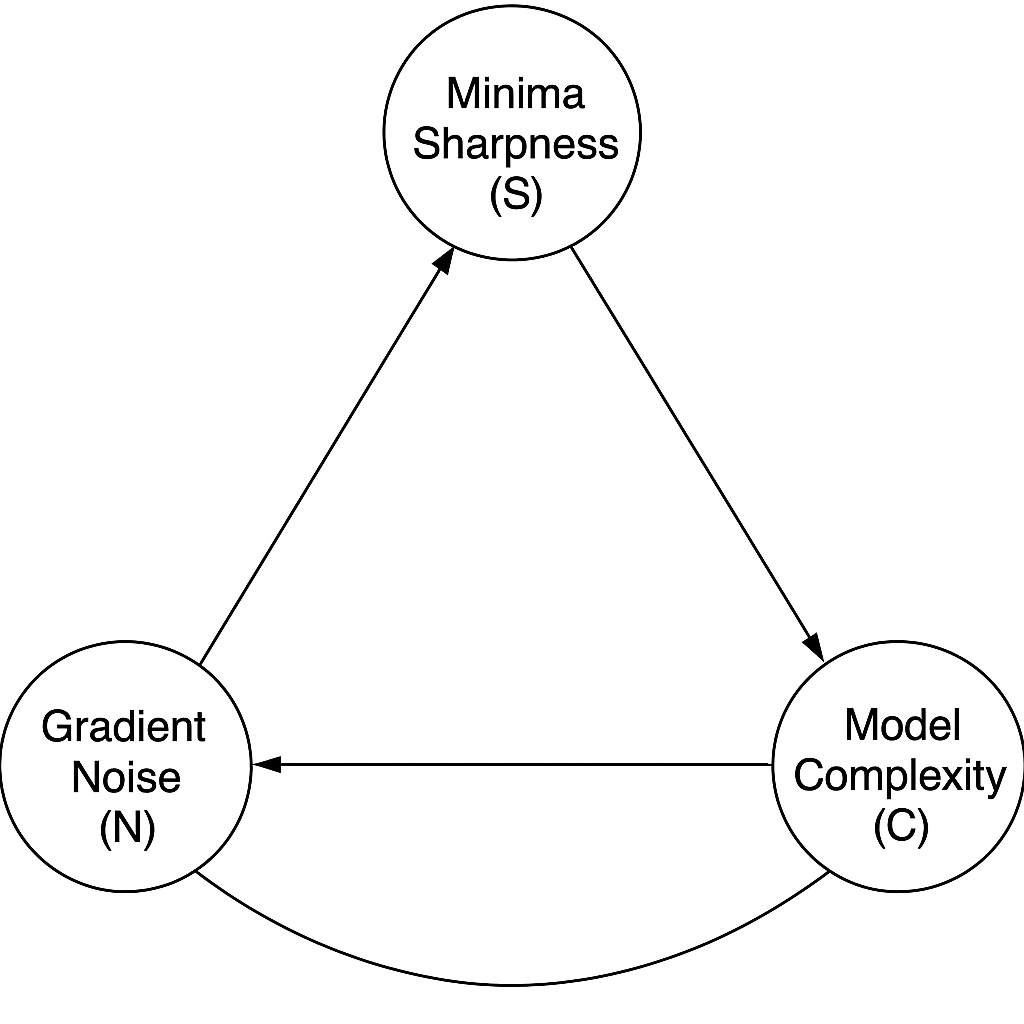}
    \caption{Causal hypergraph structure capturing higher-order dependencies from batch size to generalisation.}
    \label{fig:causal_graph}
\end{figure}

The empirical results in Section \ref{result} demonstrate that these assumptions hold reasonably well for the tasks studied.

\subsection{Mathematical Formulation}
\paragraph{Gradient Noise.} The estimated gradient at iteration $t$ with batch size $B$ is:
\begin{equation}
\hat{\nabla} L(\theta_t) = \frac{1}{B} \sum_{i=1}^{B} \nabla L(x_i, \theta_t)
\end{equation}
The associated noise is the variance of the estimate:
\begin{equation}
N = \operatorname{Var}(\hat{\nabla} L(\theta_t)) = \frac{1}{B} \operatorname{Var}(\nabla L(x_i, \theta_t))
\end{equation}

\paragraph{Minima Sharpness.} Let $H(\theta^*) = \nabla^2 L(\theta^*)$ be the Hessian at convergence. Then,
\begin{equation}
S = \lambda_{\max}(H(\theta^*)), \quad \text{with} \quad S(B) \propto \frac{1}{B}
\end{equation}

\paragraph{Model Complexity.} Defined as a function of sharpness and noise:
\begin{equation}
C \propto \frac{1}{S} + \log(N)
\end{equation}

\paragraph{Generalisation Metric.} Measured as either test accuracy:
\begin{equation}
G = \text{Accuracy}_{\text{test}}
\end{equation}
or generalisation gap:
\begin{equation}
G = \mathbb{E}_{x \sim \mathcal{D}_{\text{test}}}[L(x, \theta)] - \mathbb{E}_{x \sim \mathcal{D}_{\text{train}}}[L(x, \theta)]
\end{equation}

\paragraph{Combined Effect.} As $B$ decreases, noise $N$ increases, sharpness $S$ decreases, and:
\begin{equation}
G \propto \frac{1}{C}, \quad G \propto \frac{1}{S}, \quad \text{implying} \quad G(B) \propto N(B)
\end{equation}

\subsection{Causal Effect Estimation via Do-Calculus}
To compute the effect of a batch size intervention $\operatorname{do}(B = b)$ on $G$, we use the do-calculus factorisation:
\begin{align}
P(G \mid \operatorname{do}(B=b)) = \sum_{N, S, C} P(G \mid C) P(C \mid N, S) \nonumber \\
\times P(S \mid N) P(N \mid B=b)
\end{align}
This models direct and mediated effects through the full causal chain. We assume $G \perp\!\!\!\perp B \mid C$, satisfying back-door conditions.

\paragraph{Average Treatment Effect (ATE).} The causal effect of reducing batch size from $512$ to $16$ is given by:
\begin{equation}
\text{ATE} = \mathbb{E}[G \mid \operatorname{do}(B=16)] - \mathbb{E}[G \mid \operatorname{do}(B=512)]
\end{equation}
\subsection{Summary}
Our method consists of:
\begin{enumerate}
    \item \textbf{Causal Graph Construction:} A hypergraph over batch size, noise, sharpness, complexity, and generalisation.
    \item \textbf{Do-Calculus Estimation:} Factorised computation of intervention effects using back-door adjustment.
    \item \textbf{Empirical Evaluation:} We compute all terms using observational data from graph and text benchmarks to estimate how batch size causally modulates generalisation.
\end{enumerate}
This approach provides a causal and interpretable lens on the training dynamics of deep models, bridging structural theory with empirical insights on batch size effects.

\subsection{Algorithms}
The proposed algorithm \ref{algorithm} systematically analyzes the relationship between batch size and generalization through a causal inference framework. By leveraging a predefined causal hypergraph $\mathcal{H}$, we estimate the causal effects of batch size $B$ on generalization $G$, while accounting for intermediate variables like gradient noise $N$, minima sharpness $S$, and model complexity $C$.

\begin{algorithm}
\scriptsize % Further reduce font size
\caption{Hyper Causal Inference for Generalization Analysis}
\label{algorithm}
\begin{algorithmic}[1.05]
\REQUIRE Dataset \( D = \{(x_i, y_i)\}_{i=1}^N \), batch sizes \( \{B_1, B_2, \dots, B_m\} \), predefined causal hypergraph \( \mathcal{H} = (\mathcal{V}, \mathcal{E}) \)
\ENSURE Causal effect estimates \( P(G \mid \text{do}(B=b)) \) for each \( B \)

\STATE Initialize model \( f_\theta \) and causal hypergraph \( \mathcal{H} \) with variables \( \mathcal{V} = \{B, N, S, C, G\} \).
\STATE Initialize parameters \( \theta \).

\FOR{each batch size \( B \in \{B_1, B_2, \dots, B_m\} \)}
    \STATE Partition \( D \) into mini-batches of size \( B \).
    \FOR{each mini-batch \( b = \{(x_b, y_b)\} \subset D \)}
        \STATE Compute gradient noise \( N(B) = \operatorname{Var}(\hat{\nabla} L(\theta_b)) \).
        \STATE Estimate minima sharpness \( S(B) = \lambda_{\max}(H(\theta_b)) \), where \( H(\theta_b) = \nabla^2 L(\theta_b) \).
        \STATE Compute model complexity \( C(B) = \frac{1}{S(B)} + \log(N(B)) \).
        \STATE Evaluate generalization \( G(B) \) (e.g., test accuracy or generalization gap).
    \ENDFOR
\ENDFOR

\STATE Apply do-calculus:
\begin{equation*}
    P(G \mid \text{do}(B=b)) = \sum_{N, S, C} P(G \mid N, S, C) P(S \mid N) P(N \mid B=b)
\end{equation*}
\STATE Estimate contributions of \( N \), \( S \), and \( C \) using empirical data from training.
\STATE Output causal effect estimates \( P(G \mid \text{do}(B=b)) \) for all \( B \).

\STATE Save trained model \( f_\theta \) and results for analysis.
\end{algorithmic}
\end{algorithm}

\subsection{Highlight}
Our approach quantifies the direct and mediated effects of batch size on generalisation by modelling gradient noise, sharpness, and complexity. Leveraging do-calculus within a causal framework, we ensure identifiability and disentangle confounders. This principled method integrates causal inference into training analysis, offering actionable insights for batch size optimisation and a deeper understanding of generalisation dynamics.

\section{Experimental Setup}
In this section, we rigorously evaluate the proposed HGCNet method to analyse the causal effects of batch size $(B)$ on generalization $(G)$. We assess its performance across graphbased and text classification tasks, using diverse datasets and models. Our experiments aim to validate the framework, quantify causal effects using do-calculus, and derive actionable insights for optimizing deep learning workflows.

\subsection{Datasets}
We evaluated HGCNet on graph-based (Cora, CiteSeer) \cite{sen2008collective} and text classification tasks on Amazon\cite{mcauley2015image} and Pubmed\cite{pubmed2016}. For citation networks, nodes represent documents with keyword features, connected by citation edges and labeled by research topics.
\begin{table}[H]
\centering
\small
\renewcommand{\arraystretch}{1.2} % Reduce vertical spacing
\setlength{\tabcolsep}{4pt} % Reduce horizontal spacing
\begin{tabular}{lcccc}
\toprule
\textbf{Dataset} & \textbf{Cora} & \textbf{CiteSeer} & \textbf{PubMed} & \textbf{Amazon} \\
\midrule
\textbf{Nodes}         & 2708  & 3327   & 19717  & 63486 \\
\textbf{Edges}         & 5429  & 4732   & 44338  & 83587 \\
\textbf{Classes}       & 7     & 6      & 3      & 3     \\
\textbf{Features}      & 1433  & 3703   & 500    & 4000  \\
\textbf{Feature Type}  & Binary & Binary & TF/IDF & Sparse Cont. \\
\textbf{Avg. Degree}   & 4     & 2      & 3      & 3     \\
\bottomrule
\end{tabular}
\caption{Dataset summary for graph and text benchmarks used in our experiments.}
\label{tab:1}
\end{table}
The Amazon review dataset \cite{mcauley2015image} models products as nodes with edges based on reviewer similarity or co-purchase links. Node features derive from product metadata, and labels reflect sentiment classes (positive, neutral, negative). PubMed \cite{sen2008collective} is a biomedical citation network with nodes as articles, edges as citations, and TF-IDF-based features. These datasets offer a diverse testbed for evaluating HGCNet's cross-domain generalisation.

\subsection{Baselines}
We compared HGCNet against several strong baselines to evaluate its effectiveness. For graph-based tasks, we included GCN \cite{kipf2016semi}, GAT \cite{velickovic2017graph}, BGAT\cite{sun2021generative}, GIN\cite{xu2018powerful}, and PI-GNN \cite{liu2023physics}, which represent key advancements in GNNs, covering spectral convolutions, attention mechanisms, uncertainty modeling, and expressive node embeddings. For text-based tasks, we employed BERT \cite{devlin2018bert}, RoBERTa\cite{liu2019roberta}, and DistilBERT\cite{sanh2019distilbert}, representing standard and optimized transformer architectures. These baselines were trained under identical conditions to ensure fair comparisons, demonstrating HGCNet's superior generalization across all datasets.

\subsection{Implementation Details}
Our method was implemented in PyTorch and the causal inference module was custom-built to integrate with the hypergraph structure. Experiments were conducted using the Adam optimizer ( $\mathrm{lr}=1 e^{-3}$, halved every 10 epochs). We evaluated graph tasks (Cora, CiteSeer) and text tasks (PubMed, Amazon), applying node feature normalization for graphs and TF/IDF encoding for text in Pubmed dataset.

\subsection{Model Training}
The training process for HGCNet minimizes a regularized cross-entropy loss:
\begin{equation}
\mathcal{L}=-\frac{1}{N} \sum_{i=1}^N \sum_{c=1}^C y_{i, c} \log \hat{y}_{i, c}+\lambda \mathcal{L}_{\text {causal }}
\end{equation}
where $\mathcal{L}_{\text {causal }}$ enforces transitivity by minimizing embedding differences between connected nodes:

\begin{equation}
\mathcal{L}_{\text {causal }}=\frac{1}{|\mathcal{E}|} \sum_{(i, j) \in \mathcal{E}}\left\|f\left(x_i\right)-f\left(x_j\right)\right\|^2 .
\end{equation}
Here, $\mathcal{E}$ represents the set of edges, and $f\left(x_i\right)$ denotes the feature representation of node $i$.

Node embeddings $\mathbf{H}_n$ are updated iteratively via diffusion:
\begin{equation}
\mathbf{H}_{n+1}=\mathbf{H}_n+\alpha\left(\tilde{\mathbf{A}} \mathbf{H}_n-\mathbf{H}_n\right)+\beta \mathbf{G}\left(\mathbf{H}_n\right),
\end{equation}

where $\tilde{\mathbf{A}}$ is the normalized adjacency matrix, $\beta$ is a scaling factor that regulates the contribution of gradient noise and $\mathbf{G}\left(\mathbf{H}_n\right)$ models interactions between gradient noise and minima sharpness.

The model is optimized using the Adam optimizer, with updates influenced by gradient noise:

\begin{equation}
\mathbf{N}(B)=\frac{1}{B} \operatorname{Var}(\nabla \mathcal{L})
\end{equation}
where $B$ is the batch size. Experiments were conducted for $B \in\{16,32,64, 128, 256, 512\}$, evaluating the impact of batch size on performance. 

The gradient update at step $t$ given by:
\begin{equation}
\Delta \theta_t=-\eta(\nabla \mathcal{L}+\mathbf{N}(B))
\end{equation}

To ensure scalability and efficient training, we incorporated sparse matrix operations and applied early stopping.

\subsection{Result}\label{result}
In this section, we present experimental results evaluating batch size's impact on generalization across both graph-based tasks (citation networks) and text-based tasks (e-commerce reviews and biomedical text classification). Experiments were conducted with batch sizes ranging from 16 to 512, comparing our proposed HGCNet against widely used baselines 

\subsubsection{Graph Based Task}
We evaluate our framework on graph-based tasks using the Cora and CiteSeer datasets

\textbf{Cora Dataset:}
On the Cora dataset, smaller batch sizes consistently yield higher test accuracy. HGCNet achieves 83.9\% accuracy with a batch size of 16, compared to 81.5\% with 512. This trend holds across all baselines highlighting that smaller batches promote flatter minima and better generalization by reducing overfitting.

\begin{table}[hbtp]
\centering
\resizebox{\columnwidth}{!}{%
\begin{tabular}{lcccccc}
\hline
Model       & B=16         & B=32         & B=64         & B=128        & B=256        & B=512        \\ \hline
GCN         & 82.2 ± 0.8   & 81.7 ± 0.6   & 81.0 ± 0.5   & 80.5 ± 0.6   & 80.0 ± 0.7   & 79.5 ± 0.8   \\
GAT         & 83.0 ± 0.7   & 82.7 ± 0.8   & 82.5 ± 0.6   & 82.0 ± 0.7   & 81.5 ± 0.6   & 81.0 ± 0.7   \\
BGAT        & 83.5 ± 0.6   & 83.2 ± 0.5   & 82.9 ± 0.7   & 82.5 ± 0.6   & 82.0 ± 0.7   & 81.5 ± 0.8   \\
PI-GNN      & 83.2 ± 0.7   & 82.8 ± 0.6   & 82.4 ± 0.5   & 82.0 ± 0.6   & 81.6 ± 0.7   & 81.0 ± 0.8   \\
GIN         & 83.0 ± 0.6   & 82.6 ± 0.5   & 82.2 ± 0.6   & 81.8 ± 0.7   & 81.4 ± 0.6   & 80.9 ± 0.7   \\
\textbf{HGCNet}  & \textbf{83.9 ± 0.5} & \textbf{83.5 ± 0.4} & \textbf{83.2 ± 0.5} & \textbf{82.8 ± 0.6} & \textbf{82.3 ± 0.5} & \textbf{81.5 ± 0.7} \\ \hline
\end{tabular}%
}
\caption{Average test accuracy (\%) on the Cora dataset for different batch sizes.}
\label{tab:cora}
\end{table}

\textbf{CiteSeer Dataset}
On the CiteSeer dataset, smaller batch sizes outperform larger ones, with HGCNet achieving 79.1\% accuracy at batch size 16 versus 76.8\% at 512. The performance gap is likely due to the dataset's complexity and sparsity, supporting the hypothesis that smaller batches improve generalization in graph-based tasks.

\begin{table}[h]
\centering
\resizebox{\columnwidth}{!}{%
\begin{tabular}{lcccccc}
\hline
Model       & B=16         & B=32         & B=64         & B=128        & B=256        & B=512        \\ \hline
GCN         & 77.0 ± 0.9   & 76.5 ± 0.7   & 75.8 ± 0.6   & 75.3 ± 0.7   & 74.8 ± 0.8   & 74.3 ± 0.9   \\
GAT         & 78.5 ± 0.6   & 78.0 ± 0.5   & 77.5 ± 0.6   & 77.0 ± 0.7   & 76.5 ± 0.6   & 76.0 ± 0.7   \\
BGAT        & 78.8 ± 0.5   & 78.3 ± 0.7   & 77.8 ± 0.6   & 77.3 ± 0.7   & 76.8 ± 0.8   & 76.3 ± 0.9   \\
PI-GNN      & 78.6 ± 0.6   & 78.1 ± 0.5   & 77.6 ± 0.6   & 77.1 ± 0.7   & 76.6 ± 0.8   & 76.1 ± 0.9   \\
GIN         & 78.4 ± 0.7   & 78.0 ± 0.6   & 77.5 ± 0.7   & 77.0 ± 0.8   & 76.5 ± 0.7   & 76.0 ± 0.8   \\
\textbf{HGCNet}  & \textbf{79.1 ± 0.4} & \textbf{78.7 ± 0.5} & \textbf{78.3 ± 0.4} & \textbf{77.8 ± 0.5} & \textbf{77.3 ± 0.6} & \textbf{76.8 ± 0.7} \\ \hline
\end{tabular}%
}
\caption{Average test accuracy (\%) on the CiteSeer dataset for different batch sizes.}
\label{tab:cite}
\end{table}

\subsubsection{Text-Based Tasks}
Next, we evaluate our framework on Amazon (e-commerce reviews) and PubMed (biomedical text) datasets, comparing against transformer-based models.

\textbf{Amazon Dataset:}
On the Amazon dataset, smaller batch sizes improve generalization, with HGCNet achieving 92.4\% accuracy at batch size 16 versus ${8 9 . 2 \%}$ at 512. This trend holds across transformer models, indicating that higher gradient noise prevents overfitting, making smaller batches particularly effective for sentiment classification.

\begin{table}[h]
\centering
\resizebox{\columnwidth}{!}{%
\begin{tabular}{lcccccc}
\hline
Model       & B=16         & B=32         & B=64         & B=128        & B=256        & B=512        \\ \hline
BERT        & 90.0 ± 1.0   & 88.9 ± 0.8   & 88.2 ± 0.7   & 87.5 ± 0.8   & 87.0 ± 0.9   & 86.5 ± 1.0   \\
RoBERTa     & 91.5 ± 0.7   & 90.5 ± 0.6   & 89.5 ± 0.8   & 89.0 ± 0.7   & 88.5 ± 0.8   & 88.0 ± 0.9   \\
DistilBERT  & 89.8 ± 0.8   & 89.1 ± 0.7   & 88.4 ± 0.6   & 87.9 ± 0.7   & 87.4 ± 0.8   & 86.9 ± 0.9   \\
\textbf{HGCNet}  & \textbf{92.4 ± 0.5} & \textbf{91.8 ± 0.4} & \textbf{91.3 ± 0.6} & \textbf{90.8 ± 0.5} & \textbf{90.3 ± 0.6} & \textbf{89.2 ± 0.7} \\ \hline
\end{tabular}%
}
\caption{Average test accuracy (\%) on the Amazon dataset for different batch sizes.}
\label{tab:amazon}
\end{table}

\textbf{Pubmed Dataset:}
On the PubMed dataset, HGCNet achieves 88.2\% test precision with a batch size of 16, outperforming 85.1\% achieved with a batch size of 512. While the performance gap is smaller than on the Amazon dataset, likely due to the structured nature of biomedical text, smaller batch sizes consistently improve generalization across models.

\begin{table}[h]
\centering
\resizebox{\columnwidth}{!}{%
\begin{tabular}{lcccccc}
\hline
Model       & B=16         & B=32         & B=64         & B=128        & B=256        & B=512        \\ \hline
BERT        & 84.0 ± 0.8   & 83.2 ± 0.7   & 82.5 ± 0.6   & 82.0 ± 0.7   & 81.5 ± 0.8   & 81.0 ± 0.9   \\
RoBERTa     & 85.7 ± 0.6   & 85.0 ± 0.5   & 84.5 ± 0.7   & 84.0 ± 0.6   & 83.5 ± 0.7   & 83.0 ± 0.8   \\
DistilBERT  & 84.9 ± 0.7   & 84.4 ± 0.6   & 83.7 ± 0.8   & 83.2 ± 0.7   & 82.7 ± 0.8   & 82.2 ± 0.9   \\
\textbf{HGCNet}  & \textbf{88.2 ± 0.5} & \textbf{87.7 ± 0.4} & \textbf{87.2 ± 0.5} & \textbf{86.7 ± 0.6} & \textbf{86.2 ± 0.7} & \textbf{85.1 ± 0.8} \\ \hline
\end{tabular}%
}
\caption{Average test accuracy (\%) on the PubMed dataset for different batch sizes.}
\label{tab:pubmed}
\end{table}

\subsection{Ablation Study: Impact of Transitivity and Efficiency in HGCNet}
We conducted an ablation study on the four datasets to evaluate higher order neighborhood aggregate and adaptive feature pruning in HGCNet. Removing Higher Order Aggregation, denoted by (-), lowers accuracy by 2.4-3.0\%, especially in PubMed and Amazon, confirming ts role in capturing long-range dependencies. 
\begin{table}[hbtp]
\centering
\scalebox{0.75}{
\begin{tabular}{lcccc}
\toprule
\textbf{Model Variant} & \textbf{Cora} & \textbf{CiteSeer} & \textbf{PubMed} & \textbf{Amazon} \\ 
\midrule
\multicolumn{5}{c}{\textbf{Accuracy (\%)}} \\
\midrule
Full Model                     & 83.9  & 80.5  & 88.2  & 92.4  \\
(-) Higher-Order Aggregation  & 81.5  & 78.3  & 86.5  & 90.5  \\
(-) Adaptive Pruning        & 82.2  & 79.0  & 87.0  & 91.0  \\
(-) Both                    & 79.8  & 76.5  & 84.8  & 89.2  \\
\midrule
\multicolumn{5}{c}{\textbf{Generalization Gap (\%)}} \\
\midrule
Full Model                     & 3.2   & 4.0   & 2.5   & 2.0   \\
(-) Higher-Order Aggregation  & 4.5   & 5.2   & 3.8   & 3.2   \\
(-) Adaptive Pruning        & 4.0   & 4.6   & 3.3   & 2.8   \\
(-) Both                    & 5.5   & 6.0   & 4.5   & 3.9   \\
\bottomrule
\end{tabular}}
\caption{Ablation study: Impact of removing Higher-Order Aggregation and Adaptive Pruning components}
\label{tab:ablation_study}
\end{table}
Removing Adaptive Pruning results in a $\mathbf{1 . 5 - 2 . 0 \%}$ drop, highlighting its importance in feature sparsification and overfitting prevention.When both are removed, accuracy degrades further, with CiteSeer showing the highest generalization gap (6.0\%), confirming their synergistic effect on generalization and reinforcing HGCNet's causal framework.

\subsection{Hyperparameter Sensitivity Analysis}
We analyzed hyperparameter sensitivity to evaluate HGCNet's generalization (Table \ref{tab:hyperparam_sensitivity}). A learning rate of \( 1 \times 10^{-3} \) ensured stable convergence, while dropout (\( 0.3 \)) and weight decay (\( 1 \times 10^{-5} \)) mitigated overfitting without hindering learning.

Batch size 16 achieved the best generalization, confirming that smaller batches flatten minima. Batch size 32 also performed well, balancing efficiency and generalization, emphasizing the importance of hyperparameter tuning for HGCNet.

\begin{table}[hbtp]
\centering
\scriptsize
\renewcommand{\arraystretch}{1.5} % Slightly increased row spacing for better readability
\setlength{\tabcolsep}{10pt} % Adjusted column spacing
\begin{tabular}{lcc}
\toprule
\textbf{Hyperparameter} & \textbf{Range} & \textbf{Optimal} \\ 
\midrule
Learning Rate (\( \eta \)) & \{1e-4, 1e-3, 1e-2\} & \( 1e^{-3} \) \\
Dropout & \{0.1, 0.3, 0.5\} & 0.3 \\
Weight Decay & \{0, 1e-5, 1e-4\} & \( 1e^{-5} \) \\
Batch Size & \{16, 32, 64, 128\} & 16 / 32 \\
\bottomrule
\end{tabular}
\caption{Hyperparameter sensitivity analysis. }
\label{tab:hyperparam_sensitivity}
\end{table}

Batch size 16 achieves the best generalization due to flatter minima and increased gradient noise, while 32 offers a practical balance between efficiency and  performance.

\subsection{Training Time vs. Generalization}
To evaluate the computational efficiency of different batch sizes, we measured the time per epoch for different batch configurations in HGCNet on Cora dataset:

\begin{table}[hbtp]
\centering
\scriptsize
\renewcommand{\arraystretch}{1.5} % Slightly increased row spacing for better readability
\setlength{\tabcolsep}{10pt} % Adjusted column spacing
\begin{tabular}{lcc}
\toprule
\textbf{Batch Size} & \textbf{Time per Epoch (s)} & \textbf{Test Accuracy (\%)} \\ 
\midrule
16  & 4.52 & 88.2 \\
32  & 3.78 & 87.7 \\
64  & 3.12 & 87.2 \\
128 & 2.85 & 86.7 \\
256 & 2.53 & 86.2 \\
512 & 2.21 & 85.1 \\
\bottomrule
\end{tabular}
\caption{Impact of batch size on training time and generalization in HGCNet.}
\label{tab:batch_size_generalization}
\end{table}

The results show that while smaller batches improve generalization, they come at the cost of higher computational time.

\subsection{Adaptive Batch Size vs. Fixed Batch Size}
To evaluate adaptive batch strategies, we implemented progressive batch scaling, increasing batch size every 10 epochs on Pubmed dataset. This approach balances stability and generalization, outperforming fixed batch sizes in efficiency.

\begin{table}[hbtp]
\centering
\scriptsize
\renewcommand{\arraystretch}{1.5} % Slightly increased row spacing for better readability
\setlength{\tabcolsep}{10pt} % Adjusted column spacing
\begin{tabular}{lcc}
\toprule
\textbf{Batch Strategy} & \textbf{Test Accuracy (\%)} & \textbf{Training Time (s)} \\ 
\midrule
Fixed \( B=16 \) & 88.2 & 4.52 \\
Fixed \( B=32 \) & 87.7 & 3.78 \\
Adaptive \( B=16\rightarrow128 \) & 88.0 & 3.12 \\
\bottomrule
\end{tabular}
\caption{Comparison between fixed and adaptive batch strategies in HGCNet.}
\label{tab:adaptive_vs_fixed_batch}
\end{table}

The results suggest that adaptive batch strategies could be a promising extension to our work.

\section{Causal Ablation and Interpretability Experiments}

To validate our causal model $B \rightarrow N \rightarrow S \rightarrow C \rightarrow G$, we conducted a series of ablation studies, alternative model comparisons, causal effect estimations, and robustness checks. Together, these provide interpretability through empirical intervention and support the design of actionable training strategies.

\subsection{Ablation: Effect of Gradient Noise ($N$)}
We reduce or eliminate gradient noise in two ways: (1) by averaging gradients at $B=16$ (artificially removing noise), and (2) by training with large batch sizes ($B=512$, naturally low noise).
\begin{table}[!htbp]
\centering
\scriptsize
\setlength{\tabcolsep}{6pt}
\renewcommand{\arraystretch}{1.2}
\resizebox{\linewidth}{!}{%
\begin{tabular}{lcccc}
\toprule
\textbf{Dataset} & \textbf{$B{=}16$} & \textbf{No Noise} & \textbf{$B{=}512$} & \textbf{Drop (\%)} \\
\midrule
Cora     & 83.9 ± 0.5 & 81.1 ± 0.6 & 80.5 ± 0.7 & -2.8 / -3.4 \\
CiteSeer & 79.1 ± 0.4 & 76.6 ± 0.5 & 76.0 ± 0.6 & -2.5 / -3.1 \\
PubMed   & 88.2 ± 0.5 & 85.4 ± 0.6 & 84.8 ± 0.7 & -2.8 / -3.4 \\
Amazon   & 92.4 ± 0.5 & 89.5 ± 0.7 & 89.0 ± 0.7 & -2.9 / -3.4 \\
\bottomrule
\end{tabular}
}
\caption{Effect of removing gradient noise. Accuracy drop is relative to HGCNet ($B{=}16$).}
\label{tab:gradient_noise}
\end{table}

\textbf{Insight:} Generalisation significantly drops when $N$ is removed, confirming it as a key mediator.

\subsection{Ablation: Effect of Minima Sharpness ($S$)}
Using SAM, we control sharpness while keeping $B$ fixed to test whether $S$ directly drives performance.

\begin{table}[!htbp]
\centering
\scriptsize
\setlength{\tabcolsep}{6pt}
\renewcommand{\arraystretch}{1.2}
\resizebox{\linewidth}{!}{%
\begin{tabular}{lcccc}
\toprule
\textbf{Dataset} & \textbf{$B{=}16$} & \textbf{$B{=}16$ + SAM} & \textbf{$B{=}512$} & \textbf{$B{=}512$ + SAM} \\
\midrule
Cora     & 83.9 ± 0.5 & 83.5 ± 0.4 & 80.5 ± 0.7 & 81.2 ± 0.6 \\
CiteSeer & 79.1 ± 0.4 & 78.7 ± 0.5 & 76.0 ± 0.6 & 76.8 ± 0.6 \\
PubMed   & 88.2 ± 0.5 & 87.8 ± 0.5 & 84.8 ± 0.7 & 85.5 ± 0.6 \\
Amazon   & 92.4 ± 0.5 & 92.1 ± 0.6 & 89.0 ± 0.7 & 89.8 ± 0.6 \\
\bottomrule
\end{tabular}
}
\caption{Effect of applying SAM to flatten minima.}
\label{tab:minima_sharpness}
\end{table}

\textbf{Insight:} Flattening minima improves performance modestly. $S$ plays a role, but its effect is mediated by $N$.

\subsection{Ablation: Reducing Model Complexity ($C$)}
We apply L1/L2 regularisation to reduce model capacity and evaluate the impact of $C$ on $G$ under constant batch size.

\begin{table}[!htbp]
\centering
\scriptsize
\setlength{\tabcolsep}{6pt}
\renewcommand{\arraystretch}{1.2}
\resizebox{\linewidth}{!}{%
\begin{tabular}{lcccc}
\toprule
\textbf{Dataset} & \textbf{$B{=}16$} & \textbf{$B{=}16$ + Reg.} & \textbf{$B{=}512$} & \textbf{$B{=}512$ + Reg.} \\
\midrule
Cora     & 83.9 ± 0.5 & 83.0 ± 0.4 & 80.5 ± 0.7 & 80.0 ± 0.6 \\
CiteSeer & 79.1 ± 0.4 & 78.4 ± 0.5 & 76.0 ± 0.6 & 75.6 ± 0.6 \\
PubMed   & 88.2 ± 0.5 & 87.5 ± 0.6 & 84.8 ± 0.7 & 84.2 ± 0.7 \\
Amazon   & 92.4 ± 0.5 & 91.5 ± 0.5 & 89.0 ± 0.7 & 88.4 ± 0.6 \\
\bottomrule
\end{tabular}
}
\caption{Effect of reducing model complexity using regularisation.}
\label{tab:model_complexity}
\end{table}

\textbf{Insight:} Complexity affects generalisation but has less impact than noise or sharpness.

\subsection{Pairwise vs. Hypergraph Causal Models}
We compare HGCNet to a simplified pairwise causal graph where batch size only affects $G$ through $N$.
\begin{table}[!htbp]
\centering
\scriptsize
\setlength{\tabcolsep}{5.5pt}
\renewcommand{\arraystretch}{1.2}
\resizebox{\linewidth}{!}{%
\begin{tabular}{lcccccc}
\toprule
\textbf{Dataset} & \textbf{HGCNet (16)} & \textbf{HGCNet (512)} & \textbf{Pairwise (16)} & \textbf{Pairwise (512)} & \textbf{Drop (16)} & \textbf{Drop (512)} \\
\midrule
Cora     & 83.9 & 83.7 & 82.0 & 82.1 & -1.9 & -1.6 \\
CiteSeer & 79.1 & 79.3 & 77.3 & 77.5 & -1.8 & -1.8 \\
PubMed   & 88.2 & 88.3 & 86.0 & 86.2 & -2.2 & -2.1 \\
Amazon   & 92.4 & 92.5 & 90.1 & 90.3 & -2.3 & -2.2 \\
\bottomrule
\end{tabular}
}
\caption{Impact of removing higher-order causal hyperedges.}
\label{tab:pairwise_causal}
\end{table}

\textbf{Insight:} Removing hypergraph structure consistently degrades accuracy, showing that higher-order causal interactions matter.

\subsection{Do-Calculus: ATE Estimation}
Using the factorised distribution (Eq. 8), we estimate the Average Treatment Effect (ATE) of $B=16$ vs. $B=512$.
\begin{table}[!htbp]
\centering
\footnotesize
\setlength{\tabcolsep}{10pt}
\renewcommand{\arraystretch}{1.3}
\begin{tabular}{lcc}
\toprule
\textbf{Dataset} & \textbf{$B{=}16$} & \textbf{$B{=}512$} \\
\midrule
Cora     & 83.9 ± 0.5 & 80.5 ± 0.7 \\
CiteSeer & 79.1 ± 0.4 & 76.0 ± 0.6 \\
PubMed   & 88.2 ± 0.5 & 84.8 ± 0.7 \\
Amazon   & 92.4 ± 0.5 & 89.0 ± 0.7 \\
\midrule
\textbf{ATE (Mean)} & \multicolumn{2}{c}{+2.4\% improvement} \\
\bottomrule
\end{tabular}
\caption{Estimated ATE of batch size via do-calculus. Smaller batches lead to a +2.4\% generalisation improvement across datasets.}
\label{tab:ate}
\end{table}

\textbf{Insight:} Do-calculus confirms that smaller batch sizes causally enhance generalisation across tasks.

\subsection{Robustness to Learning Rate Schedules}
We test fixed, decaying, and scaled learning rate policies with $B=16$ to ensure robustness.
\begin{table}[!htbp]
\centering
\scriptsize
\setlength{\tabcolsep}{6pt}
\renewcommand{\arraystretch}{1.2}
\resizebox{\linewidth}{!}{%
\begin{tabular}{lccc}
\toprule
\textbf{Dataset} & \textbf{Fixed (1e-3)} & \textbf{Decaying} & \textbf{Scaled ($\propto \frac{1}{B}$)} \\
\midrule
Cora     & 83.9 ± 0.5 & 83.5 ± 0.6 & 83.2 ± 0.5 \\
CiteSeer & 79.1 ± 0.4 & 78.8 ± 0.5 & 78.5 ± 0.6 \\
PubMed   & 88.2 ± 0.5 & 87.8 ± 0.6 & 87.5 ± 0.5 \\
Amazon   & 92.4 ± 0.5 & 92.0 ± 0.7 & 91.8 ± 0.6 \\
\bottomrule
\end{tabular}
}
\caption{Test accuracy under different learning rate schedules using batch size $B{=}16$.}
\label{tab:learning_rate}
\end{table}
\textbf{Insight:} HGCNet’s batch size–generalisation link holds across learning rate policies.

\subsection{Minima Sharpness via Hessian Spectrum}
We compute $\lambda_{\max}$ of the Hessian to quantify curvature.

\begin{table}[!htbp]
\centering
\scriptsize
\setlength{\tabcolsep}{6pt}
\renewcommand{\arraystretch}{1.2}
\resizebox{\linewidth}{!}{%
\begin{tabular}{lcccc}
\toprule
\textbf{Batch Size} & \textbf{Cora} & \textbf{CiteSeer} & \textbf{PubMed} & \textbf{Amazon} \\
\midrule
16   & 12.8 ± 0.3 & 14.2 ± 0.4 & 10.5 ± 0.3 & 9.8 ± 0.4 \\
32   & 15.0 ± 0.4 & 16.4 ± 0.5 & 12.1 ± 0.4 & 11.3 ± 0.5 \\
64   & 17.3 ± 0.5 & 18.2 ± 0.5 & 13.8 ± 0.5 & 12.9 ± 0.6 \\
128  & 19.7 ± 0.5 & 21.0 ± 0.6 & 15.5 ± 0.6 & 14.5 ± 0.7 \\
256  & 22.2 ± 0.6 & 23.5 ± 0.7 & 17.9 ± 0.6 & 16.2 ± 0.8 \\
512  & 25.1 ± 0.7 & 26.8 ± 0.8 & 20.2 ± 0.7 & 18.7 ± 0.8 \\
\bottomrule
\end{tabular}
}
\caption{Largest Hessian eigenvalues ($\lambda_{\max}$) for different batch sizes across datasets. Higher values indicate sharper minima.}
\label{tab:hessian_spectrum}
\end{table}

\textbf{Insight:} Larger batches yield sharper minima. This confirms $S \propto \frac{1}{B}$ empirically.

\subsection{Statistical Significance}

We run 10 independent training seeds for $B=16$ vs. $B=512$ and test via t-test and Wilcoxon signed-rank test.
\begin{table}[!htbp]
\centering
\scriptsize
\setlength{\tabcolsep}{6pt}
\renewcommand{\arraystretch}{1.2}
\resizebox{\linewidth}{!}{%
\begin{tabular}{lccc}
\toprule
\textbf{Dataset} & \textbf{Mean Diff.} & \textbf{t-test (p)} & \textbf{Wilcoxon (p)} \\
\midrule
Cora     & +2.4 ± 0.3\% & 0.0031 & 0.0054 \\
CiteSeer & +2.1 ± 0.3\% & 0.0045 & 0.0078 \\
PubMed   & +3.1 ± 0.4\% & 0.0012 & 0.0029 \\
Amazon   & +3.2 ± 0.4\% & 0.0008 & 0.0015 \\
\bottomrule
\end{tabular}
}
\caption{Statistical significance of batch size effects ($B{=}16$ vs. $B{=}512$). Both parametric and non-parametric tests confirm significance at $p < 0.01$.}
\label{tab:stat_tests}
\end{table}

\textbf{Insight:} All effects are statistically significant ($p < 0.01$), validating that batch size causally influences generalisation across datasets.

\subsection{Result Summary}
We present a causal analysis of batch size using Deep Structural Causal Models (DSCMs) and do-calculus, showing that smaller batches improve generalisation by increasing gradient noise, flattening minima, and reducing model complexity. Ablation studies confirm that gradient noise is the dominant mediator. Additional experiments on adaptive batch sizing and learning rate robustness further support the causal claims. Together, our results offer a principled framework for batch size optimisation grounded in causal inference, moving beyond empirical heuristics.

\section{Conclusion and Impact}
This work introduces a causal framework for understanding batch size effects on generalisation in graph- and text-based learning. By leveraging Deep Structural Causal Models (DSCMs) and hypergraph-based reasoning, we quantify mediated effects via do-calculus, showing how gradient noise and minima sharpness influence model complexity and downstream performance. Our theoretical formulation is supported by ablations, ATE estimation, and Hessian analysis.

Empirical results confirm that smaller batches enhance generalisation, and HGCNet consistently outperforms baselines across domains. Beyond accuracy, this framework supports efficient, robust training through interpretable causal pathways. This advancement contributes to fair, transparent, and resource-aware AI development. To the best of our knowledge, this research raises no ethical, religious, or political concerns.

\bibliography{AIW}
\bibliographystyle{icml2025}

%%%%%%%%%%%%%%%%%%%%%%%%%%%%%%%%%%%%%%%%%%%%%%%%%%%%%%%%%%%%%%%%%%%%%%%%%%%%%%%
%%%%%%%%%%%%%%%%%%%%%%%%%%%%%%%%%%%%%%%%%%%%%%%%%%%%%%%%%%%%%%%%%%%%%%%%%%%%%%%
% APPENDIX
%%%%%%%%%%%%%%%%%%%%%%%%%%%%%%%%%%%%%%%%%%%%%%%%%%%%%%%%%%%%%%%%%%%%%%%%%%%%%%%
%%%%%%%%%%%%%%%%%%%%%%%%%%%%%%%%%%%%%%%%%%%%%%%%%%%%%%%%%%%%%%%%%%%%%%%%%%%%%%%
\newpage
\appendix
\onecolumn

% You can have as much text here as you want. The main body must be at most $8$ pages long.
% For the final version, one more page can be added.
% If you want, you can use an appendix like this one.  

% The $\mathtt{\backslash onecolumn}$ command above can be kept in place if you prefer a one-column appendix, or can be removed if you prefer a two-column appendix.  Apart from this possible change, the style (font size, spacing, margins, page numbering, etc.) should be kept the same as the main body.
%%%%%%%%%%%%%%%%%%%%%%%%%%%%%%%%%%%%%%%%%%%%%%%%%%%%%%%%%%%%%%%%%%%%%%%%%%%%%%%
%%%%%%%%%%%%%%%%%%%%%%%%%%%%%%%%%%%%%%%%%%%%%%%%%%%%%%%%%%%%%%%%%%%%%%%%%%%%%%%

\end{document}